\documentclass[sn-mathphys-num,iicol]{sn-jnl}

\usepackage{graphicx}%
\usepackage{multirow}%
\usepackage{amsmath,amssymb,amsfonts}%
\usepackage{amsthm}%
\usepackage{mathrsfs}%
\usepackage[title]{appendix}%
\usepackage{xcolor}%
\usepackage{textcomp}%
\usepackage{manyfoot}%
\usepackage{booktabs}%
\usepackage{algorithm}%
\usepackage{algorithmicx}%
\usepackage{algpseudocode}%
\usepackage{listings}%

\theoremstyle{thmstyleone}%
\theoremstyle{thmstyletwo}%

\theoremstyle{thmstylethree}%

\begin{document}

\title[Article Title]{Efficient Pretraining Model based on Multi-Scale Local Visual Field Feature Reconstruction for PCB CT Image Element Segmentation}


\author{\fnm{Chen} \sur{Chen}}

\author{\fnm{Kai} \sur{Qiao}}

\author{\fnm{Jie} \sur{Yang}}

\author{\fnm{Jian} \sur{Chen}}

\author{\fnm{Bin} \sur{Yan*}}


\abstract{Element segmentation is a key step in nondestructive testing of Printed Circuit Boards (PCB) based on Computed Tomography (CT) technology. In recent years, the rapid development of self-supervised pretraining technology can obtain general image features without labeled samples, and then use a small amount of labeled samples to solve downstream tasks, which has a good potential in PCB element segmentation. At present, Masked Image Modeling (MIM) pretraining model has been initially applied in PCB CT image element segmentation. However, due to the small and regular size of PCB elements such as vias, wires, and pads, the global visual field has redundancy for a single element reconstruction, which may damage the performance of the model. Based on this issue, we propose an efficient pretraining model based on multi-scale local visual field feature reconstruction for PCB CT image element segmentation (EMLR-seg). In this model, the teacher-guided MIM pretraining model is introduced into PCB CT image element segmentation for the first time, and a multi-scale local visual field extraction (MVE) module is proposed to reduce redundancy by focusing on local visual fields. At the same time, a simple 4-Transformer-blocks decoder is used. Experiments show that EMLR-seg can achieve 88.6\% mIoU on the PCB CT image dataset we proposed, which exceeds 1.2\% by the baseline model, and the training time is reduced by 29.6 hours, a reduction of 17.4\% under the same experimental condition, which reflects the advantage of EMLR-seg in terms of performance and efficiency.}

\keywords{Multi-scale local visual field, PCB nondestructive testing, Mask image modeling, Semantic segmentation}



\maketitle

\section{Introduction}\label{sec1}
Printed circuit boards are important components of electronic equipment. Their manufacturing quality affects the stability of electronic products. In order to ensure normal operation of equipment, PCB non-destructive testing using CT imaging technology is a very important step in its production and failure analysis. Among the main elements of PCB, vias are channels for connecting wires on both sides of a circuit board, wires are the basis for connecting circuit components, and pads are the basic building blocks of surface mount components. Whether these three types of elements can be accurately identified is directly related to the subsequent steps of defect testing. Therefore, the element segmentation is the key link of non-destructive testing.

Applying image processing to PCB CT image element segmentation are commonly used methods, which mainly include traditional image processing algorithms\cite{bib1,bib2} and supervised semantic segmentation models based on deep learning\cite{bib3,bib4,bib5,bib6}. Although these methods are effective, traditional image processing algorithms are difficult to handle a large number of complex image features and generally can only achieve single-target segmentation; supervised models based on deep learning require a large number of labels to achieve better segmentation effects, but the pixel-level labels required for segmentation cost a lot of manpower annotation.

With the rapid development of self-supervised learning, the self-supervised pretraining and finetuning paradigm is widely used in semantic segmentation. This paradigm utilizes large-scale pretraining models to learn general image representations and then outperforms supervised model segmentation using only a small amount of labeled data for finetuning. Masked image modeling (MIM)\cite{bib7} is one of the pretraining methods. This method divides the image into non-overlapping image patches. After randomly masking some patches, the model uses the remaining visible patches to learn general image representations and then reconstructs the original image. The reconstruction target can be pixels\cite{bib8,bib9} or certain features of the original image\cite{bib7, bib10,bib11,bib12}, or image feature generated by the online teacher network\cite{bib13,bib14,bib15,bib16,bib17}, such as iBOT\cite{bib13} and dBOT\cite{bib14}. 

Inspired by self-supervised learning, CD-MAE\cite{bib18} proposed by Song et al. uses MIM as a pretraining method to achieve segmentation of wires, vias and pads, verifying the effectiveness of MIM in PCB CT image element segmentation. However, CD-MAE does not consider the difference between PCB CT images and natural images. For an image, the global visual field refers to the entire image, and the local visual field refers to part of the image. Different from natural images, the area of elements in PCB CT images accounts for a small proportion of the whole image and elements usually has a regular shape. Most of the elements and connection relationship can be reconstructed through the semantic of a local visual field, and image representation can be fully learned in the process. On the contrary, the global view has a large amount of semantic redundancy for PCB element reconstruction, which may decrease pretraining efficiency and even damage model performance. Therefore, if we can highlight the semantic information of local visual fields of reasonable sizes, it may enable the pretraining model to learn more efficiently and achieve a balance between efficiency and performance.

Based on the analysis above, we propose an efficient pretraining model based on multi-scale local visual field feature reconstruction for PCB CT image element segmentation (EMLR-seg). This model introduces the teacher-guided mask image modeling pretraining model into the PCB CT image element segmentation for the first time. Compared with reconstructing original image pixels, the reconstruction target generated by teacher network can reduce the influence of low-level semantic information (such as color, geometry, texture, etc.) on reconstruction. And in order to eliminate redundancy from global visual field , We propose the multi-scale local visual field extraction (MVE) module. In the vanilla Transformer model, A 224×224 image is divided into 14×14 non-overlapping image patches. In order to take into account different sizes of PCB elements, we choose the local field with 5 × 5, 7 × 7, 9 × 9 image patches. Since MVE forces the model to efficiently utilize local visual field information during reconstruction, it can accelerate the convergence and improve the performance on PCB CT image element segmentation. Experiments show that EMLR-seg can achieve 88.6\% mIoU on the PCB CT image dataset we proposed, which exceeds 1.2\% of the baseline model, and the training time is reduced by 29.6 hours under the same experimental condition, which reflects the advantage of EMLR-seg in terms of performance and efficiency.
EMLR-seg has following contributions to PCB CT image element segmentation:
(1) EMLR-seg introduces teacher-guided MIM pretraining method into PCB CT image element segmentation for the first time, which reduces the influence of pixel-level semantic information.
(2) In view of the unique characteristics of PCB CT images, we introduce MVE module to efficiently utilize image information. MVE accelerates model's convergence and improves performance.
(3) It has been experimentally verified that the effect of 4 Transformer blocks as a decoder can achieve a balance between efficiency and performance.

\section{Related work}\label{sec2}
\subsection{PCB CT image element segmentation}\label{subsec1}
In PCB defect detection, the inspection of elements and components is a key step, and the semantic segmentation algorithm provides a new way for the detection of elements and components. In 2015, Song et al.\cite{bib2} proposed an automatic segmentation algorithm for vias in PCB CT images based on superpixel algorithm. In 2017, Qiao et al.\cite{bib3} proposed a wire detection algorithm using Fully Convolutional Network. In 2020, Ulbert et al. used Mask R-CNN to implement via detection, and Li et al.\cite{bib4} used constructed random forest pixel classifier to perform semantic segmentation of PCB components. In addition, there are some methods that use traditional algorithms or deep learning methods to segment PCB components\cite{bib5,bib6}. These methods only detect a single type of element or segment components. That is because pixel-level annotation of components and elements is very labor-intensive, therefore the size of the dataset is limited. And the number of model parameters is small, resulting in weak feature learning capabilities. In addition, Song et al. proposed CD-MAE\cite{bib18}, which uses the self-supervised pretraining and finetuning method to realize the segmentation of three elements: wires, vias and pads, but CD-MAE does not consider the difference between PCB CT image and natural image, and simply combines MAE\cite{bib8} with contrast learning.

\subsection{Semantic segmentation using deep learning}\label{subsec2}
Semantic segmentation aims to assign predefined semantic categories to each pixel in the image. The use of deep learning for semantic segmentation is mainly divided into supervised and unsupervised methods. Supervised method refers to using labeled data to train the model and adjust its own parameters by comparing the differences between its output and the real label. This method mainly goes through the following stages of development: using encoder-decoder architecture, which recovering characteristic dimensions and spatial information of the images through the decoder\cite{bib19,bib20,bib21}; using multi-scale target fusion to improve feature extraction\cite{bib22}; optimizing the convolution algorithm to expand the extraction receptive field in the convolution scale\cite{bib23,bib24}; introduce transformer architecture to obtain superior global modeling capabilities\cite{bib25,bib26,bib27}; build large multi-modal datasets and use interactive annotation to achieve high-precision semantic segmentation\cite{bib28}.
Since the supervised method requires pixel-level labels, it requires a lot of manpower and training time, and has limitations in practical applications.

The self-supervised method refers to using a large amount of unlabeled data for self-supervised pretraining to obtain an encoder that can extract general features of images, and then using a small amount of labeled data to finetune on semantic segmentation. Classified according to the pretraining model task, these methods mainly include generative methods and contrastive ones. The generative methods learn feature representation through images reconstruction\cite{bib7}. The contrastive methods use individual similarities and differences in images to train the model, and uses the clustering idea to train through a dual-branch architecture\cite{bib29,bib30,bib31,bib32,bib33}. Since there are multiple elements in PCB CT images, while contrastive models are usually more effective for single-target image classification problems, we choose to use MIM method which performs well and belongs generative models.

\subsection{Mask image modeling(MIM)}\label{subsec3}
MIM is one of the most important methods in self-supervised pretraining. This method first divides the original image into patches, then masks a certain proportion of patches, learning image representation and reconstructing images through the encoder and decoder. The reconstructed target signals are diverse. Initially, the original image pixels are generally used, such as MAE\cite{bib8}. Other reconstruction targets include discrete tokens\cite{bib7,bib10,bib12}, histogram features of gradients (HOG)\cite{bib11}, and features extracted by synchronous training teacher network\cite{bib13,bib14} etc. At present, the last type can achieve the best results in nature image for downstream tasks, so we use dBOT\cite{bib14}, which belongs to the last type, as the baseline model.

dBOT adopts a teacher-student network architecture. The student completely adopts MAE structure and is responsible for reconstruction. The reconstructed target signal is generated by the teacher network, which only has an encoder. Multiple breakpoints are set in the training. The parameters of student network are updated to teacher network at breakpoints and are not updated in other periods. Compared with pixel, the reconstructed target signal generated by the teacher network reduces the impact of low-level semantics such as color and texture. It is more suitable for PCB CT images with metal artifacts, uneven grayscale, translucency and other noises.

\section{Method}\label{sec3}

\subsection{Pretraining stage}\label{subsec4}
\begin{figure}[h]%
\centering
\centerline{\includegraphics[width=0.55\textwidth]{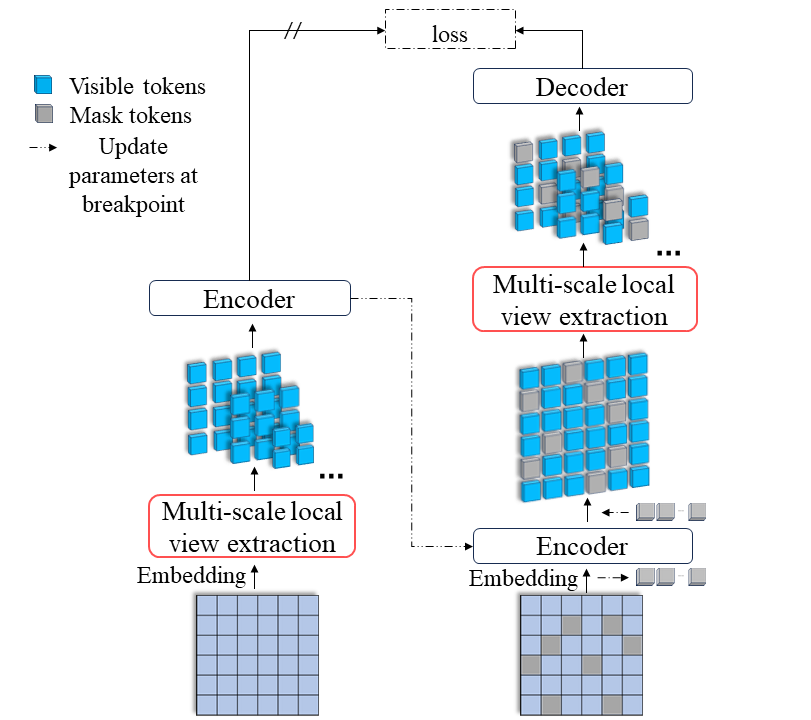}}
\caption{Our EMLR-seg architecture. The student network consists of an encoder, a MVE module and a decoder. The teacher network consists of a MVE module and an encoder.}\label{fig1}
\vspace{-0.2cm}
\end{figure}

Our pretraining model adopts a MIM model with a teacher-student network architecture. As shown in Figure~\ref{fig1}, the teacher network generates the reconstruction target signal, and the student network performs reconstruction task. The training goal is to obtain a student network encoder that can obtain general features of images.

\subsubsection{Student Network: Perform reconstruction task}\label{subsubsec1}

The student network uses MAE’s asymmetric encoder-decoder architecture, and adopts the Transformer block as the standard layer. The default size of the input image is 224×224, and the size of each image patch is 16×16, so a complete image has a total of 14×14 image patches. 

\textbf{Encoder. }The encoder $ \mathrm{E}_{s} $ contains 24 Transformer blocks.  As in the standard vanilla ViT operation, the patched image is represented by $ x_{s} $, and then embedded by a linear projection with added positional embeddings $ p_{s}^{v} $. Then a certain proportion of patches (for example, 60\%) is masked, and only the remaining visible patches $ x_{s}^{v} $ is input to the encoder. The output is encoded visible patch features $ f_{s}^{v} $. The mask operation can reduce the calculation amount of the encoder. The encoder can be represented by the formula:
\begin{equation}
f_{s}^{v}=\mathrm{E}_{s}\left(x_{s}^{v}+p_{s}^{v}\right)
\label{eq1}
\end{equation}

\textbf{Multi-scale local visual field extraction module (MVE).}In order to avoid the negative impact of redundant information and enable the model to focus on a reasonable local visual field during reconstruction, we add a MVE module before performing reconstruction task. 

\begin{figure}[h]%
\centering
\vspace{-0.2cm}
\centerline{\includegraphics[width=0.35\textwidth]{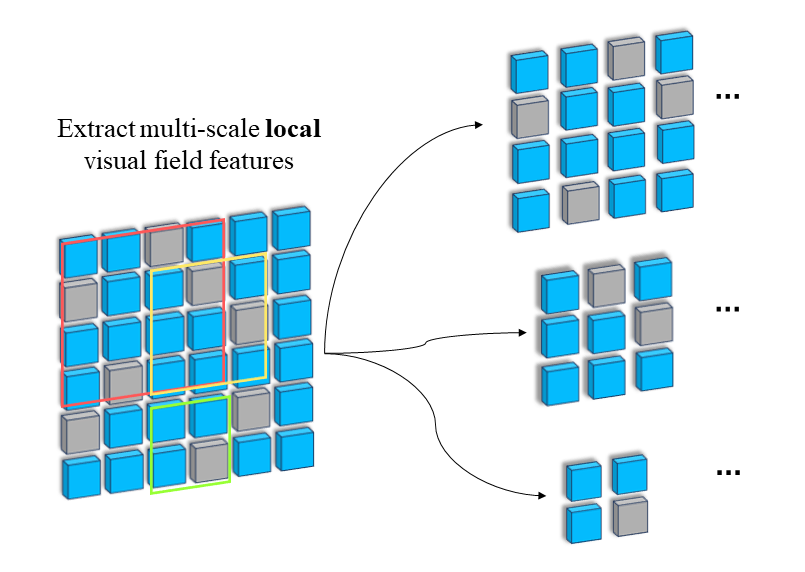}}
\vspace{-0.2cm}
\caption{Multi-scale local visual field extraction module in the student network. Firstly, the encoder's output and the mask tokens are combined and arranged in the order of the original image. Then the local visual field features are extracted respectively as reconstruction units. }\label{fig2}
\vspace{-0.2cm}
\end{figure}

The input of this module is all encoded visible patches $ f_{s}^{v} $ and mask tokens $ f_{s}^{m} $ ,and each mask token is a learned vector that represents a missing patch. Firstly, they are arranged according to the initial position of the original image, and then features are extracted according to the local visual field size of 5×5, 7×7, 9×9 patches respectively. In order to ensure that the number of patches covered by different sizes of local visual fields are roughly the same, we assign different amounts for them, which is 4, 2, 1 respectively. Each extracted part serves as a separate reconstruction unit. Then the module’s output features are fed into the decoder for local reconstruction.  

In order to ensure that the teacher network and the student network act on the same local visual field, it is also necessary to record the position information extracted by MVE. MVE can be represented by the formula:

\begin{equation}
f_{s}^{m l}=\operatorname{MVE}\left(f_{s}^{v}+f_{s}^{m}\right)
\label{eq2}
\end{equation}

where $ f_{s}^{m l} $ represents all local visual field features.
 When calculating attention matrices of a Transformer block, extracting the local visual field can reduce the computational complexity of matrices. Assuming that the image has a total of image patches, the time complexity when calculating an attention matrix of Transformer is $ O\left(h^{2} w^{2}\right) $.  EMLR-seg selects the local visual field, assuming that the size of the selected window is $ l_{i} \times l_{i} $, then the time complexity is $ O\left(\sum l_{i}^{4}\right) $. If $ h \times w \gg l_{i} \times l_{i} $, the time complexity of calculating attention matrices would be significantly reduced.

\textbf{Decoder. }The decoder $ \Gamma_{s} $ contains a total of 4 Transformer blocks. The task of the decoder is to reconstruct the reconstruction target generated by the teacher network. We take visual fields extracted by MVE as separate units, and input them into the decoder to predict the reconstruction target. Since redundant information is reduced after extracting the local visual field, a simple decoder should be used. We choose a simple combination of 4 ViT blocks as the decoder. Ablation experiments show that a suitable encoder can improve model performance. The decoder can be represented by the formula:
\begin{equation}
F_{s}^{m l}=\Gamma_{s}\left( f_{s}^{m l}\right)
\label{eq3}
\end{equation}
where $ F_{s}^{m l} $ represents student’s output feature.
\vspace{-0.2cm}

\subsubsection{Teacher Network: Generate a reconstruction target}\label{subsubsec2}

The teacher network is responsible for generating reconstruction targets. The complete image patches $ x_{t}^{m l} $ is directly input directly into MVE. In order to ensure that the teacher network and the student network act on the same local visual field, the same local area is selected in the complete image through position information extracted by MVE. Then the local visual field tokens are fed into a 24-Transformer-block encoder $ \mathrm{E}_{t} $. In addition, the operations of linear projection and adding positional embedding $ p_{t}^{m l} $ are the same as those of the student network, but the teacher network doesn’t update the gradient during training. The teacher network can be represented by the formula:

\begin{equation}
F_{t}^{m l}=\mathrm{E}_{t}\left(x_{t}^{m l}+p_{t}^{m l}\right)
\label{eq4}
\end{equation}
where $ F_{t}^{m l} $ represents teacher’s output feature.

\textbf{Update parameters at the training breakpoint. }The training is divided into two stages, and a breakpoint is set in the middle of two stages. In the first stage, the teacher network encoder adopts randomly initialized parameters. At the breakpoint, all parameters of the student network encoder are copied to the teacher network and used as the initialization parameters of teacher in the second stage, providing a richer visual representation than the first stage. The teacher’s output is used as the reconstructed target signal to optimize the student network.

Following the optimization approach of dBOT, we adopt the representation generated by the teacher network as the reconstruction target for MIM. We adopt SmoothL1loss as the loss function and only calculate the loss of mask patches.
\vspace{-0.2cm}
\begin{equation}
	\operatorname{Loss}_{r e c}=\frac{1}{R} \sum^{R}\left|F_{t}^{m l(\text { mask })}-F_{s}^{m l(\text { mask })}\right|
\label{eq5}
\end{equation}

where $ F_{t}^{m l} $ represents teacher’s output feature, $ F_{s}^{m l} $ represents student’s output feature, and $ R $ represents the sum of multi-scale visual field feature.

\subsection{Finetuning stage}\label{subsec5}

\begin{figure}[h]%

\centering
\hspace{-0.35cm}
\includegraphics[width=0.5\textwidth]{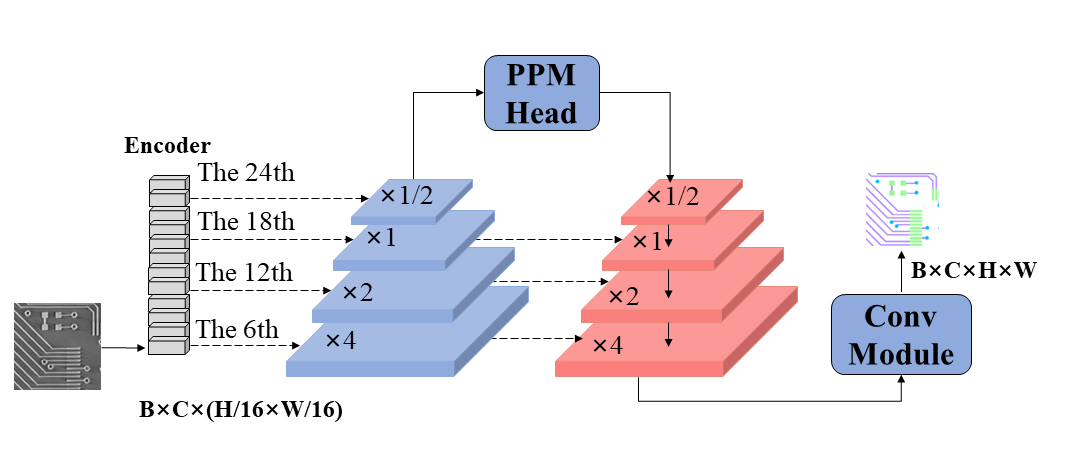}

\caption{The finetuning model. Where B, C, H, W respectively represent the batch size, the channel, image height and width.}\label{fig3}

\end{figure}

After pretraining, the student encoder has obtained general representation of images through image reconstruction. In order to achieve semantic segmentation, the decoder need to be replaced by a segmentation head in the finetuning stage. In the current image pretraining and finetuning paradigm, UperNet\cite{bib34} is generally used as the segmentation head. UperNet is a unified perception parsing network.As shown in Figure~\ref{fig3}, we combine the pretrained encoder with UperNet to form a finetuning network. Firstly, the features of the 6th, 12th, 18th, and 24th layers of the encoder are extracted and changed to 1/4, 1/8, 1/16, and 1/32 of the original image through dimension transformation. These features are input into the feature pyramid network(FPN) module of UperNet, and after feature fusion, the segmentation results are output.

We use the cross-entropy loss function to calculate the loss between segmentation results and the labels, as shown in the formula:

\vspace{-0.45cm}
\begin{equation}
\begin{split}
\operatorname{Loss}_{\text {seg }}
&=\operatorname{CrossEntropyLoss}\left(I_{\text {pred}}, I_{g t}\right)\\
\end{split}
\label{eq6}
\end{equation}

where $ I_{\text {pred }} $ represents EMLR-seg segmentation result and $ I_{g t} $ represents the original accurate label, namely the ground truth.

\section{Experiments}\label{sec4}
\subsection{Experiment setup}\label{subsec6}

\begin{figure}[h]%
\vspace{-0.2cm}
\centering
\includegraphics[width=0.45\textwidth]{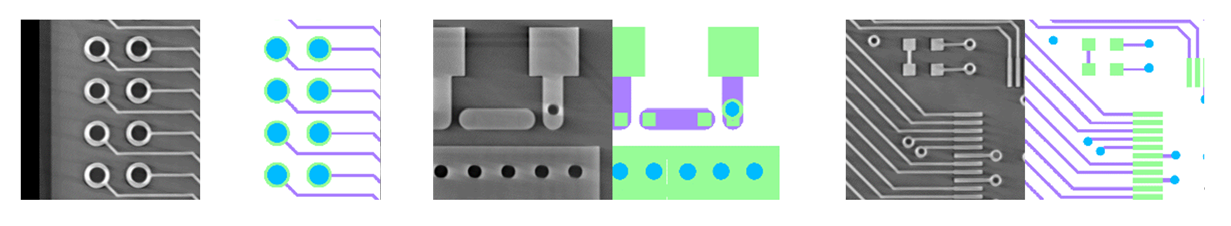}
\caption{Examples of PCB CT images and labels.  The blue pixels are vias, the green ones are pads, and the purple ones are wires.}\label{fig4}
\end{figure}

\textbf{Datasets. }Our dataset is divided into two parts. One is the unlabeled dataset used for pretraining. The number of images reaches 400k and the resolution is 512×512. Fine-tuning stage uses a labeled dataset, with 2366 samples in training set, 500 samples in test set and 718 samples in valuation set, and the resolution is 512×512. The annotation data contains three types of elements, namely vias, wires and pads. Image examples are shown in Figure~\ref{fig4}.

\textbf{Default setting. }Our experiment is carried out on 4 Tesla V100 DGXS. The image input size is 224×224. In the pretraining phase, we use ViT-L as the backbone. The student network contains a 24-Transformer-blocks encoder, and a 4-Transformer-blocks decoder. The teacher network only contains a 24-Transformer-blocks encoder and the gradient is not updated during training. The first training stage is 80 epochs, and the second one is 220 epochs. In the comparative experiments, we use 400k unlabeled images for pretraining, and other experiments use 50k images. We use random clipping with scale range from 0.9 to 1, random horizontal flipping and normalization as data enhancement. By default, we use AdamW optimizer for training with 40 warmup epochs. The initial learning rate is $10^{-3}$, and the weight decay is 0.05. In the finetuning stage, the number of training epoch is 40 with 5 warmup epochs.

\textbf{Evaluation Metrics. }In the semantic segmentation task of natural images, a commonly used performance evaluation metrics is mean Intersection over Union (mIoU). The calculation method of mIoU is: the intersection of the predicted area and the actual area of one element divided by the union of the two. In this way, the single-category IoU is calculated, and the average value of all types of elements is mIoU. We believe that PCB CT image element segmentation can also use this evaluation metrics. The mIoU calculation method is shown in the formula:

\vspace{-0.4cm}

\begin{equation}
\begin{split}
m I o U&=\frac{1}{k+1} \sum_{i=0}^{k} \frac{G T \cap \operatorname{Pr} e d}{G T \cup \operatorname{Pr} e d}\\
\end{split}
\label{eq7}
\end{equation}

where $ k $ represents the number of categories. $ G T \cap Pred $ represents the intersection between the ground truth and the prediction result, and $ G T \cup Pred $ represents the union of them.

\subsection{Comparative experiments}\label{subsec7}

In Table~\ref{tab1}, we conducted PCB CT image element segmentation experiments on commonly used semantic segmentation models. Supervised models include models based on CNN and Transformer, and self-supervised models include MAE and dBOT. The labeled data used by all models are exactly the same, and self-supervised models use the same segmentation head for finetuning. As shown in the table, the segmentation performance of our model is significantly better than supervised methods, and the performance of the baseline model is further improved by EMLR-seg, which shows that our analysis of PCB CT image characteristics in the introduction is accurate. 

\begin{table}[h]
\vspace{-0.6cm}
\caption{Comparison of PCB CT image element segmentation performance between some supervised models and self-supervised models}
	\centering
	\begin{tabular}{lcc}
		\toprule
		                                       & Model           & mIoU(\%) \\ \cmidrule{2-3}
		\multirow{6}{*}{Supervised model}      & U-Net           & 79.1     \\
		                                       & U-2-Net         & 84.6     \\
		                                       & PSPnet          & 82.6     \\
		                                       & Deeplab v3+     & 84.3     \\
		                                       & SETR            & 76.2     \\
		                                       & Segformer       & 82.3     \\ \midrule
		\multirow{5}{*}{Self-supervised model} & MAE(ViT-L)      & 86.7     \\
		                                       & CD-MAE(ViT-L)   & 87.5     \\
		                                       & dBOT(ViT-L)     & 87.4     \\
		                                       & EMLR-seg(ViT-L) & \textbf{88.6}     \\ \bottomrule
	\end{tabular}
	\label{tab1}
 \vspace{-0.2cm}
\end{table}

\begin{figure}[h]%
\centering
\hspace{-0.35cm}
\includegraphics[width=0.45\textwidth]{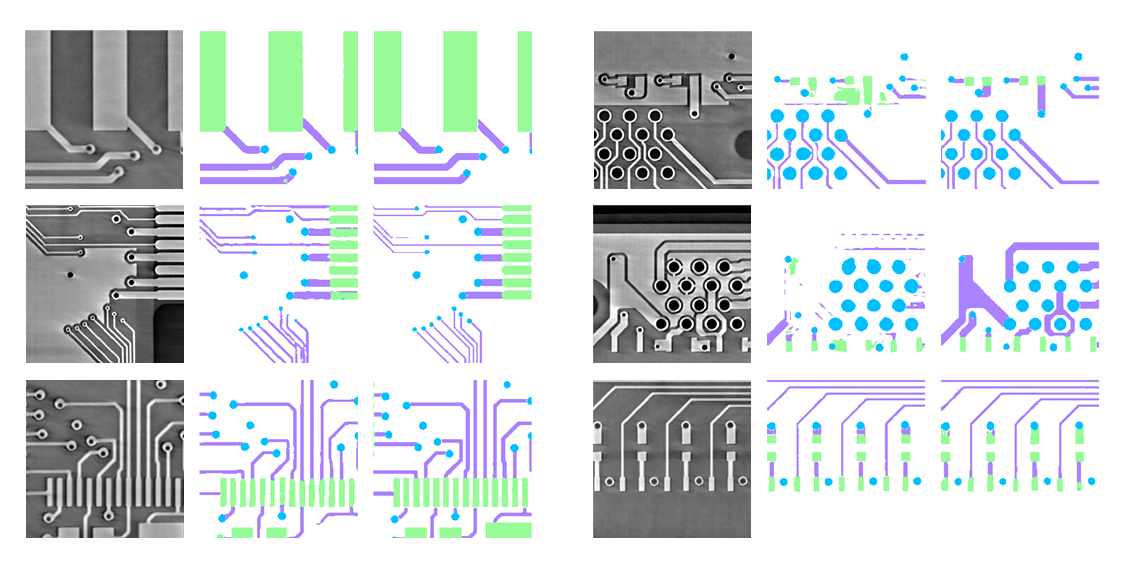}
\caption{Visualization results of element segmentation. The three columns are the original image, dBOT and EMLR-seg segmentation results in turn.}\label{fig5}
\end{figure}

As shown in Figure~\ref{fig5}, the segmentation accuracy of EMLR-seg is better than dBOT. Especially the ability to distinguish the boundaries of wires and pads has been significantly improved, and the edges of elements are more regular.

\subsection{Efficiency experiment}\label{subsec8}
Table~\ref{tab2} compares segmentation performance with different training epochs of dBOT and EMLR-seg. It can be seen that the baseline model dBOT achieves the best mIoU of 87.4\% when trained for 169.7 hours (400 epoches), while EMLR-seg can achieve the best mIoU of 88.6\% when trained for 140.1 hours (300 epoches), 29.6 hours lesser than dBOT. At the same time, EMLR-seg is close to the best mIoU of dBOT at 200 epoches. This shows that EMLR-seg not only improves the segmentation performance of PCB CT image elements, but also improves the training efficiency, verifying the analysis of PCB CT image element characteristics in the introduction.

\begin{table}[h]
\vspace{-0.6cm}
\caption{Segmentation performance comparison of different training epochs with dBOT and EMLR-seg}
 \setlength{\tabcolsep}{12pt}
	\centering
	\begin{tabular}{lccc}
		\toprule
		Model    & Epoch & Time (h) & mIoU(\%) \\ 
  \midrule
		dBOT     & 300   & 127.3    & 86.8      \\
		dBOT     & 400   & 169.7    & 87.4      \\
		EMLR-seg & 200   & 93.4     & 87.3      \\
		EMLR-seg & 250   & 116.7    & 88.0      \\
		EMLR-seg & 300   & 140.1    & \textbf{88.6}      \\
		EMLR-seg & 400   & 186.7    & \textbf{88.6}      \\ 
  \bottomrule
	\end{tabular}
	\label{tab2}
\vspace{-0.2cm}
\end{table}

\subsection{Main property experiments}\label{subsec9}

\textbf{Decoder depth. }The student network decoder is responsible for reconstruction task, so its structure is very important. If the structure is too complex, it will compensate the encoder function to learn image feature representation, which will affect the finetuning result. Otherwise, the reconstruction task may not be completed. The baseline model uses 8 ViT blocks. We carry out ablation experiments on decoder depth, and result in Table~\ref{tab3} shows that 4 blocks can achieve the best result, which also proves that extracting local vision can reduce information redundancy and improve training efficiency.

\begin{table}[h]
\setlength{\tabcolsep}{10pt}
\vspace{-0.1cm}
\vspace{-0.2cm}
\centering
\caption{Segmentation performance comparison of different decoder depths}\label{tab3}%
\begin{tabular}{@{}lccc@{}}
\toprule
Decoder depth & 2  & 4  & 8 \\
\midrule
mIoU(\%)    & 86.6   & \textbf{87.4}  & 86.2  \\
\botrule
\end{tabular}
\vspace{-0.2cm}
\end{table}

\begin{figure}[h]%
\vspace{-0.4cm}
\centering
\hspace{-0.5cm}
\centerline{\includegraphics[width=0.45\textwidth]{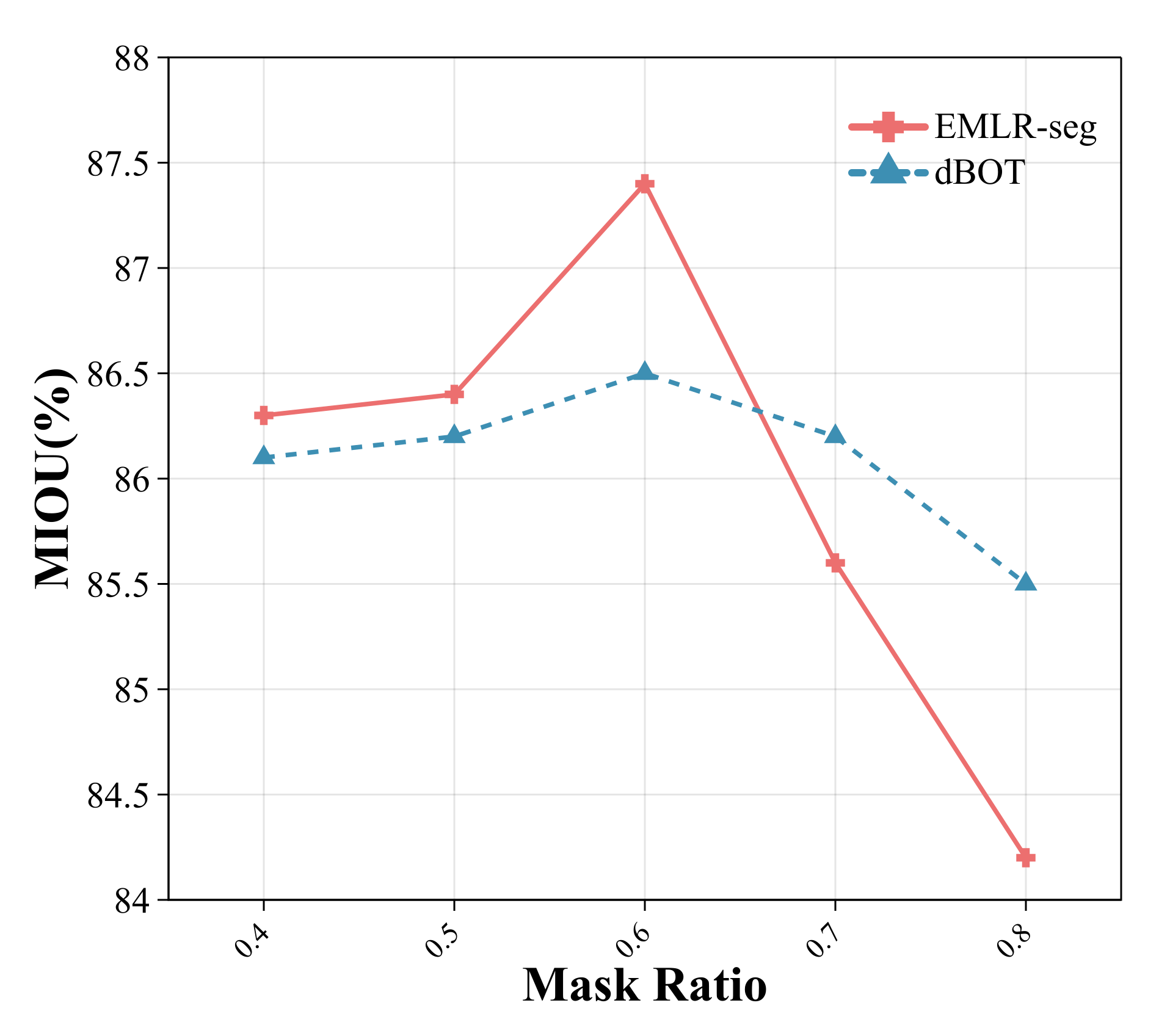}}
\vspace{-0.2cm}
\caption{Segmentation performance comparison of different mask ratios}\label{fig6}
\end{figure}

\textbf{Mask ratio. }According to MAE’s explanation of MIM, in the process of extracting general features of images by the model, there is a large amount of redundant information in an image. By masking a large proportion of the image, redundant information can be eliminated, and the difficulty of image reconstruction will be increased, forcing the model to learn more accurate image features. Figure~\ref{fig6} shows the segmentation performance after pretraining using different mask ratios. It can be seen that PCB CT images achieve the best result with a mask ratio of 60\%. When the mask ratio is low, the segmentation accuracy decreases slightly due to the influence of redundant information. When the mask ratio is too high, too much mask results in insufficient information for reconstruction. This problem is more obvious in local visual fields than global visual field. Therefore, compared to dBOT, the performance of EMLR-seg drops significantly.

\begin{figure}[h]%
\centering
\vspace{-0.4cm}
\centerline{\includegraphics[width=0.5\textwidth]{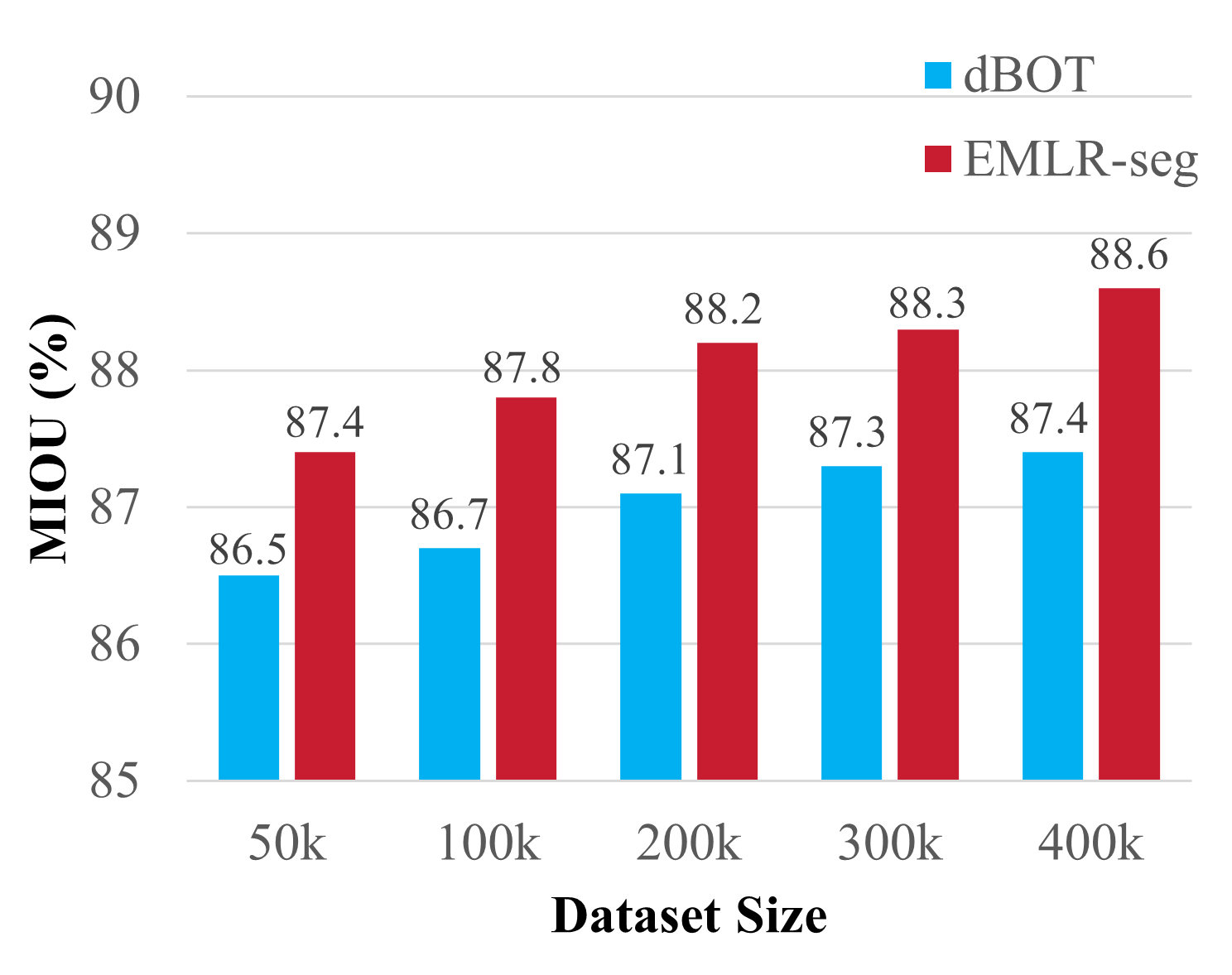}}
\vspace{-0.4cm}
\caption{Segmentation performance comparison of different dataset sizes}\label{fig7}
\end{figure}

\textbf{Dataset size. }As shown in Figure~\ref{fig7}, we test segmentation performance under different dataset sizes and find that EMLR-seg outperforms the baseline model on all dataset sizes. As data increases, model performance continues to improve, achieving 88.6\% mIoU on the dataset of 400 thousand PCB CT images, and the speed does not slow down, indicating that there is still possibility of improving performance as data increases.

\textbf{Local visual field size. }We compared the segmentation results using different local visual fields in the pretraining stage. As shown in Table~\ref{tab4} (experiments using 50 thousands images), when using a single size of local visual field, the best performance can be obtained by 7 × 7 patches. And multi-scale performance is better than a single-scale, which proves the effectiveness of MVE module.

\begin{table}[h]
\vspace{-0.2cm}
\centering
\caption{Segmentation results of reconstruction using different local visual fields}
\begin{tabular}{lcccc}
\toprule
Visual field size & 5×5  & 7×7  & 9×9  & \textbf{Multi-scale} \\ 
\midrule
mIoU(\%)          & 86.7 & 87.1 & 86.9 & \textbf{87.4}       \\
  \botrule
	\end{tabular}
	\label{tab4}
 \vspace{-0.2cm}
\end{table}

\vspace{-0.3cm}

\section{Conclusion and analysis}\label{sec5}
In this paper, in order to make the self-supervised pretraining model more suitable for PCB CT element segmentation task, we introduce the teacher-guided mask image reconstruction pretraining model into the task for the first time, greatly reducing the impact of image noise on feature learning. In view of the small and regular characteristics of PCB CT image elements, we propose MVE module, which eliminates the global information redundancy for a single PCB CT element. It not only improves pretraining efficiency, but also improves the performance in element segmentation. 

At present, we use the full finetuning method, which finetunes all parameters of the model. In the future we will try to use the Parameter-Efficient Finetuning to improve finetuning efficiency. At the same time, the problems such as the imbalance of categories and the difficulty of edge segmentation can’t be solved in pretraining, so we will strive to achieve it by adjusting the finetuning model architecture later.


\bibliography{sn-bibliography}

\end{document}